# Beyond Proximity: A Keypoint-Trajectory Framework for Classifying Affiliative and Agonistic Social Networks in Dairy Cattle

**Sibi Parivendan [1], Kashfia Sailunaz [2] and Suresh Neethirajan [1,2]***
[1] Faculty of Computer Science, Dalhousie University, 6050 University Avenue, Halifax, NS B3H 4R2, Canada
[2] Faculty of Agriculture, Dalhousie University, Truro, NS B3H 4R2, Canada
* Correspondence: sneethir@gmail.com

**Abstract**
Precision livestock farming requires objective assessment of social behavior to support herd welfare monitoring, yet most existing approaches infer interactions using static proximity thresholds that cannot distinguish affiliative from agonistic behaviors in complex barn environments. This limitation constrains the interpretability and usefulness of automated social network analysis in commercial settings. In this study, we present a pose-based computational framework for interaction classification that moves beyond proximity heuristics by modeling the spatiotemporal geometry of anatomical keypoints. Rather than relying on pixel-level appearance or simple distance measures, the proposed method encodes interaction-specific motion signatures from keypoint trajectories, enabling differentiation of social interaction valence. The framework is implemented as an end-to-end computer vision pipeline integrating YOLOv11 for object detection (mAP@0.50: 96.24%), supervised individual identification (98.24% accuracy), ByteTrack for multi-object tracking (81.96% accuracy), ZebraPose for 27-point anatomical keypoint estimation, and a support vector machine classifier trained on pose-derived distance dynamics. On annotated interaction clips collected from a commercial dairy barn, the interaction classifier achieved 77.51% accuracy in distinguishing affiliative and agonistic behaviors using pose information alone. Comparative evaluation against a proximity-only baseline demonstrates that keypoint-trajectory features substantially improve behavioral discrimination, particularly for affiliative interactions. The results establish a proof-of-concept for automated, vision-based inference of social interactions suitable for constructing interaction-aware social networks. The modular architecture and low computational overhead support near-real-time processing on commodity hardware, providing a scalable methodological foundation for future precision livestock welfare monitoring systems.

## 1. Introduction

Understanding dairy cow behavior is essential for effective herd management, animal welfare improvement, and productivity enhancement in commercial dairy operations. Recognition of fundamental behaviors such as lying, standing, walking, and feeding, alongside more complex patterns including lameness detection, gait analysis, and social interactions, enables early disease identification and proactive welfare interventions. Timely detection and management of health problems like lameness minimize long-term negative impacts on productivity, reproductive performance, and animal well-being [1]. Beyond individual health, social interactions among cows, whether affiliative such as grooming and allogrooming, or agonistic such as headbutting and displacement, offer critical insight into stress levels, dominance hierarchies, resource access patterns, and overall herd cohesion [2]. Comprehensive monitoring of these behaviors supports holistic herd welfare assessment and data-driven decision-making in precision livestock management.

Behavior monitoring in dairy cattle has traditionally relied on contact sensors including accelerometers, magnetometers, and radio frequency identification ear tagging systems [3, 4]. These sensor-based systems, while accurate for specific behaviors, encounter significant practical limitations such as animal discomfort, sensor detachment or loss, high maintenance costs, and limited scalability across large herds. Although earlier generations of three-dimensional vision and wearable sensing systems showed promising improvements in locomotion tracking and calving detection, device durability and data integration challenges persisted in farm-level applications [5]. This limitation catalyzed a transition toward non-invasive computer vision methods, which have gained substantial momentum in automated animal motion and behavior recognition systems [6, 7].

Recent advances in machine learning (ML) and deep learning (DL) models integrated with computer vision have transformed livestock behavior analysis. Deep neural networks such as C3D, ConvLSTM, and DenseNet-based spatiotemporal architectures achieve 86 to 98% accuracies in classifying single-animal behaviors including walking, lying, and feeding [8, 9]. However, pixel-level models impose substantial computational demands, requiring long video sequences for reliable inference while exhibiting reduced robustness in visually complex barn environments where occlusion and crowding are endemic [10, 11]. The field has thus recognized that advancing precision livestock farming requires computational methods that overcome these pixel-level constraints.

Menezes et al. [12] recently highlighted the integration of artificial intelligence (AI) and large language models (LLMs) in livestock monitoring, demonstrating how visual data enable continuous welfare assessment. The ongoing digital transformation of livestock farming through AI, biometric recognition, and sensor fusion has been explored extensively [13]. Furthermore, Yang et al. [14] demonstrated that computer vision now supports objective body conformation evaluation and management optimization in dairy farms. These developments collectively represent a paradigm shift from sensor-based data acquisition to vision-based behavioral analytics, enabling noninvasive monitoring of large herds while reducing labor costs and observer bias.

Skeleton-based action recognition has emerged as a computationally efficient alternative to pixel-level methods. Pose estimation models extract structured representations of posture and movement through anatomical keypoint detection, providing lightweight and interpretable features that reduce reliance on massive datasets [15, 16]. Three-dimensional vision systems for cattle growth monitoring have been comprehensively reviewed, encompassing 47 relevant studies [17] emphasize their importance for advancing automated livestock management. Li et al. [18] developed a hybrid RGB-skeleton fusion model for lameness detection achieving 97% accuracy with only 2.65% errors under challenging barn lighting conditions. However, the computational complexity and training overhead of multimodal fusion approaches present barriers to practical deployment. These challenges have motivated the evolution of keypoint-based modeling as a central computational tool in digital livestock analytics.

Keypoint detection has proven particularly valuable for individual animal identification in herds where coat pattern recognition proves ineffective. Menezes et al. [19] achieved 96% identification accuracy using Euclidean distances between anatomical keypoints and Siamese neural networks, maintaining high performance even when new cows entered the dataset. Deep learning pose estimation models have successfully detected locomotion abnormalities and health conditions across species, with de Paula et al. [20] demonstrating robust performance in sow locomotion tracking for early disease detection. These frameworks establish the technical foundation for precision livestock systems capable of continuous, objective welfare assessment.

Despite significant progress in single-animal behavior recognition, a critical gap persists: automated analysis of social interactions for constructing behavioral social networks remains underexplored. Most empirical studies still infer interactions through proximity thresholds, classifying two cows as interacting if they remain within a fixed distance for a defined duration [7]. This approach conflates fundamentally distinct behavioral phenomena, affiliative bonding and agonistic conflict, into a single undifferentiated signal. Consequently, the inability to capture interaction valence severely limits understanding of group dynamics, social stress, and herd welfare [21].

Social network analysis (SNA) has emerged as an essential methodological framework for quantifying social structures and predicting behavioral relationships in livestock systems. Marina et al. [2] demonstrated that SNA applied to real-time location data accurately predicts social behavior in dairy cattle, revealing how parity, kinship, and spatial proximity influence social bond strength. Participatory SNA [22] has been applied to examine collective action and social cohesion in agricultural communities. Parivendan et al. [23] reviewed the integration of SNA and DL for precision dairy monitoring, identifying the critical requirement for methods capable of automatically distinguishing affiliative and agonistic behaviors from continuous video. Nevertheless, current empirical implementations depend largely on manually annotated or sensor-derived data, severely limiting temporal granularity and scalability for commercial farm deployment.

This study directly addresses this recognized gap by introducing a modular, end-to-end computer vision system for automated social network analysis of dairy cattle from standard barn video footage. The principal innovation is a novel interaction inference methodology that transcends proximity-based heuristics by analyzing temporal trajectories of anatomical landmarks to classify dyadic interactions. Rather than detecting co-location alone, this framework models the geometric and dynamic signatures of interactions, enabling principled separation of affiliative and agonistic behaviors through a computational approach fundamentally distinct from existing methods.

The proposed pipeline integrates five architectural components: (i) cow detection, (ii) individual cow identification, (iii) continuous tracking, (iv) anatomical keypoint detection, and (v) interaction classification via temporal keypoint analysis. The pipeline implements YOLOv11 for cow detection and identification [24, 25], ByteTrack for multi-object tracking [26], ZebraPose for keypoint detection [27], and support vector machine (SVM) classification for behavioral inference. This integrated architecture enables automatic conversion of raw video into structured interaction data suitable for network construction. The design synthesizes lessons from prior sensor-based systems that successfully identified locomotion and calving events, extending these principles to non-contact video analytics [28].

The proposed system was trained on naturally occurring barn interactions recorded under real farm conditions, ensuring ecological validity and authentic social dynamics. The framework targets three interaction categories: headbutting, displacement, and licking or grooming that comprehensively capture both positive and negative social behaviors. Related multi-class architectures have been explored in animal behavior modeling, such as AnimalFormer, which employs multimodal vision transformers for interaction analysis [29. The modular design and low computational footprint enable deployment on edge computing devices, supporting real-time monitoring and social network analysis in commercial barn settings [30].

The two novel contributions of this research are:

1. Temporal keypoint-based interaction classification: This approach decodes behavioral valence from skeletal motion patterns, replacing conventional reliance on extensive spatial features extracted from entire video sequences.
2. Efficient candidate selection via proximity-based filtering: This method identifies cow pairs warranting detailed keypoint analysis using object detection bounding boxes to detect proximity for durations exceeding three seconds, thereby substantially reducing computational overhead while maintaining behavioral accuracy.

In contrast to behavioral descriptive studies, this research represents a genuine methodological advance in digital livestock systems. By integrating pose-based motion analysis with graph-theoretic social network construction, the framework establishes a computational foundation for scalable, objective, and noninvasive welfare monitoring aligned with precision livestock farming principles.

## 2. Methodology

### *2.1. Study Site and Recording Setup*

A custom dataset was used in this study that was collected at a commercial dairy farm in Sussex, New Brunswick, Canada on May 5, 2025. This farm was organized into 2 primary sections: a milking section with 60 cows, and a dry section with 5 dry cows. There were also intermittent observations of a sick dry cow. All the cows in both sections were unconfined and moved about freely, which facilitated the observation of their behaviors in a quiet and natural state. Furthermore, the lack of visual identifiers, ear tags, or any other wearable equipment represented that inter-cow interactions were free of any external interference.

The data collection setup included five GoPro Hero13 Black cameras with Ultrawide Lens recording in 4K at 60 fps. For the primary setup, three cameras were mounted dorsally on ceiling-mounted bird cages, while the fourth camera was mounted on the top of the dry cow section, and the fifth one was on a tripod at a lateral viewpoint surrounding the milking area. All cameras collected continuous videos between 10:00 AM and 6:00 PM on that same day, collecting about seven hours of footage. The GoPro's internal stabilizing improved footage quality by eliminating mount shakes, and cuts wind disturbance. The lighting was almost consistent throughout the duration of the data collection, and the recorded footage covered the entire barn including functional zones such as feeding stations, milking areas, brushes and rest areas. This contributed to achieving higher recorded spatial interactions.

### *2.2. Dataset Preparation and Annotation*

This study presents the methodological framework and validation on a focused dataset from the dry cow section to establish reproducible benchmarks for the interaction inference module. The modular architecture is designed for generalization across different barn configurations, herd compositions, and recording conditions, as demonstrated through the integration of external benchmark datasets (A36 Cows) in the keypoint detection module training. Future work will extend validation across the full barn dataset and multi-farm deployments. Total 375 images were extracted for the training dataset which consisted of frames selected from 2 videos from the dry cow section and each video had a continuous presence of multiple cows. The first video duration was 39 minutes and 30 seconds and the second video was 35 minutes and 14 seconds long. Frames were extracted every 12 seconds to ensure there was sufficient diversity of positions. Each of the sampled images showed 3 to 6 cows, averaging about 5 cows per frame.

#### 2.2.1 Object Detection

The 375 extracted images were manually annotated using the CVAT annotation tool [31], resulting in a total of 1930 labeled bounding box instances. Each bounding box contained one cow contributing to the foundation of the object detection performance evaluation. Additionally, the bounding boxes were pivotal in the initial phases of cow localization.

#### 2.2.2. Object Re-Identification

For object re-identification (i.e., identifying each cow individually), the cropped images of individual cows were extracted from the annotated bounding boxes and were then sorted into 6 identity classes based on visual attributes. IDs were designated for the 5 dry cows and one sick cow present in the dry cow section during data collection. Finally, 1,889 cropped images were stored with 355, 286, 392, 454, 374, and 28 images per class respectively. The lowest image count (i.e., 28) belonged to the sick cow. This dataset was subsequently used to train cow re-identification models.

#### 2.2.3. Keypoint Detection

27 anatomical keypoints were detected from the limbs, face, spine, and tail of each cow using the 27-point anatomical model of ZebraPose [27]. CVAT was used again to aid in the manual annotation process, which was completed by two separate annotators. Some variation in spatial labeling was observed, especially in partially occluded regions such as face and limbs. To assess inter-annotator agreement, a subset of 200 randomly selected cow images was jointly annotated and evaluated. Cohen's Kappa coefficient for anatomical keypoint placement across these images was 0.88, indicating strong agreement [32]. The marked disagreements were resolved through consensus review, and ambiguous frames were excluded from training to ensure label quality. A total of 1,956 cows were annotated, each with 27 points, amounting to over 51,000 annotations, of which some were occluded [33].

#### 2.2.4. Interaction Inference

An additional 160 video clips were handpicked from the same dry cow section for the interaction annotation that displayed pairs of cows interacting. Each clip was labeled as one of the three classes- affiliative interactions (licking and grooming), agonistic headbutting, or displacement. Additionally, continuous keypoint trajectories for both animals from the interactions were included in the input set for the interaction classifier. The video clips durations varied between 6 to 15 seconds.

### *2.3. Computer Vision Pipeline Overview*

An end-to-end computer vision pipeline is proposed in this paper to derive social interactions among dairy cows from video footage captured in the barn using object detection, individual cow re-identification, multi-object tracking, anatomical keypoint detection, and interaction classification. The output of each stage is used as input to the next, which enables the automated cow-to-cow interaction monitoring in a temporally consistent manner over long periods of unstructured video footage. The overall workflow is demonstrated in Figure 1. This integrated pipeline provides behavioral quantifications in a fully automated manner in a naturalistic, commercial barn setting and in doing so, allows for the identification of cow interactions with minimal manual intervention. Each component of the pipeline is described step-by-step in the following subsections.

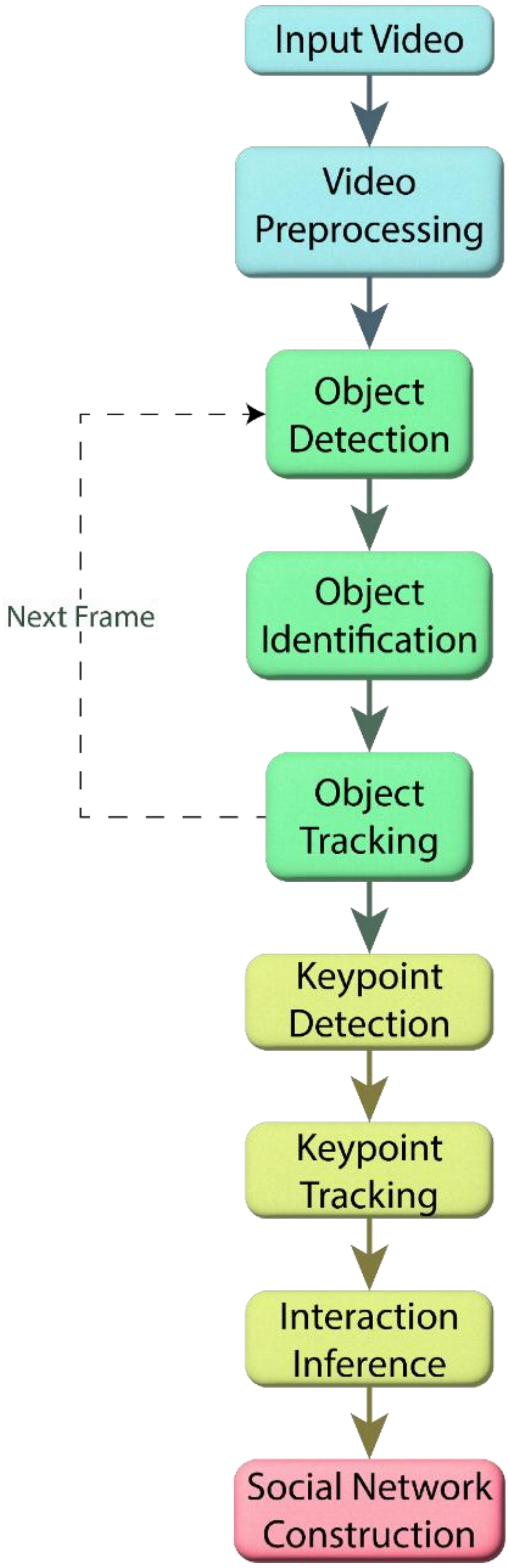

**Figure 1.** Overview of the complete pipeline for automated behavioral inference of dairy cattle, integrating object detection, identification, multi-object tracking, keypoint estimation, and interaction inference.

*2.4. Object Detection*

Accurate localization of cows in video frames forms the backbone of the pipeline, as all subsequent processes, including tracking, identification, and behavior analysis, are built on the initial bounding box detections. For this task, YOLOv11x was used, which is a high-performance, state-of-the-art real-time single-shot object detector [34]. It is an ideal choice for systems that continuously capture video, such as barn surveillance, due to its high speed and accuracy.

The YOLOv11x model was trained on the COCO dataset [35], which has over 118,000 annotated images in 80 object categories, including cows. This facilitated large scale training, which enabled the model to learn generalizable and transferrable features for detecting animals in diverse visual setting, including different cow postures, orientations, and occlusions. The manually annotated 375 images from the dry cow section with the bounding boxes were used for barn-specific evaluation. This helped to assess the model's performance in generalization to barn environments, including occlusions, overlapping individuals, and bedding clutter. Barn-specific fine-tuning was implemented afterwards using the custom dataset to improve model precision.

The fine-tuning phase used the following hyperparameter configuration: the model was trained for 100 epochs with a batch size of 16 and an image resolution of 640×640. The learning rate started at 0.01 and decayed to 0.01, with a momentum of 0.937 and weight decay of 0.0005. A 3-epoch warm-up was applied. Data augmentation included horizontal flipping (fliplr = 0.5), scaling (scale = 0.5), color jittering (hsv_s = 0.7), translation (translate = 0.1), erasing (erasing = 0.4), and RandAugment. Training was optimized using Automatic mixed precision (AMP) and deterministic execution for reproducibility and efficiency.

This hyperparameter configuration was selected to balance real-time processing requirements with detection accuracy, achieving near-real-time performance suitable for continuous farm monitoring. The pipeline sequentially employed YOLOv11x to analyze each frame of the video and produced a bounding box for each detected cow. These detections were then immediately passed to the object re-identification and multi-object tracking module.

### *2.5. Object Re-Identification and Multi-Object Tracking*

Consistent tracking of individual animals over time is essential for behavioral modeling and inferring social interactions. This requires not only detection of the cows in each frame, but also the identification of the same cow through different frames, and tracking it over time. In this paper, this task was completed in two steps: (i) object re-identification using a classification model trained on cropped cow images; and (ii) multi-object tracking using an appearance and motion based tracker.

#### 2.5.1. Object Re-Identification

In order to achieve cow-level identity consistency, the classification variant YOLOv11x architecture was adapted to recognize different cows with respect to their physical appearances. A dataset of 1,930 manually cropped images with six different cows from the dry cow section was prepared with no additional data augmentation. The cropped images preserved the original resolution and included identifying markers that were crucial for visual identity recognition such as fur patterns, horns, ears, and relative size. The classification dataset was divided into training, validation, and testing sets with 70:20:10 ratio. YOLOv11x-cls model used the default hyperparameters and included a classification head efficient for small sample, few class learning.

Specifically, the model was trained for 120 epochs with a batch size of 64 and an input resolution of 224×224 pixels. The optimization settings included a learning rate of 0.01, momentum of 0.937, and weight decay of 0.0005. RandAugment and erasing (erasing = 0.4) were used to improve robustness, alongside horizontal flipping (fliplr = 0.5) to account for lateral pose variability. The model was trained using AMP (automatic mixed precision) and deterministic mode, and ran on a single-GPU setup (device=0). These configurations enabled accurate identity labeling even in the presence of pose variation and occlusions. Finally, the predicted identities for each bounding box were employed for assigning cow-specific track IDs throughout the entire workflow. This consistent identity tracking served as the basis for mapping interactions in pairs of cows.

### 2.5.2. Multi-Object Tracking

In order to achieve continuity from one frame to the next, the primary algorithm employed for multi-object tracking was ByteTrack [26]. As a strong tracking-by-detection approach, ByteTrack assigns and associates detected objects over a series of frames by a combination of Kalman filtering and Intersection over Union (IoU) based matching. Unlike appearance-dependent trackers, ByteTrack predicts the motion of objects, allowing it to work effectively within dense crowds. This is necessary within the context of animal tracking, as they can often look quite similar and their tracking paths can overlap to the point of causing a collision.

In this phase, the cow tracking was performed using the bounding boxes created by the YOLOv11x detector, and the identity labels from the re-identification model were used to reinforce long-term track consistency. Three specific hyperparameter tuning were necessary for barn conditions- (i) ‘track threshold’ to control confidence level for new object initialization; (ii) ‘match threshold’ for determining spatial overlap between the predicted and detected boxes; and (iii) ‘track buffer’ to maintain temporary track records during brief occlusions or detection gaps. The result consisted of continuous movement paths for every individual cow in the video sequence. The subsequent modules utilized these paths and the corresponding locations for spatial and behavioral analysis. Figure 2 shows a sample of the cow tracking process.

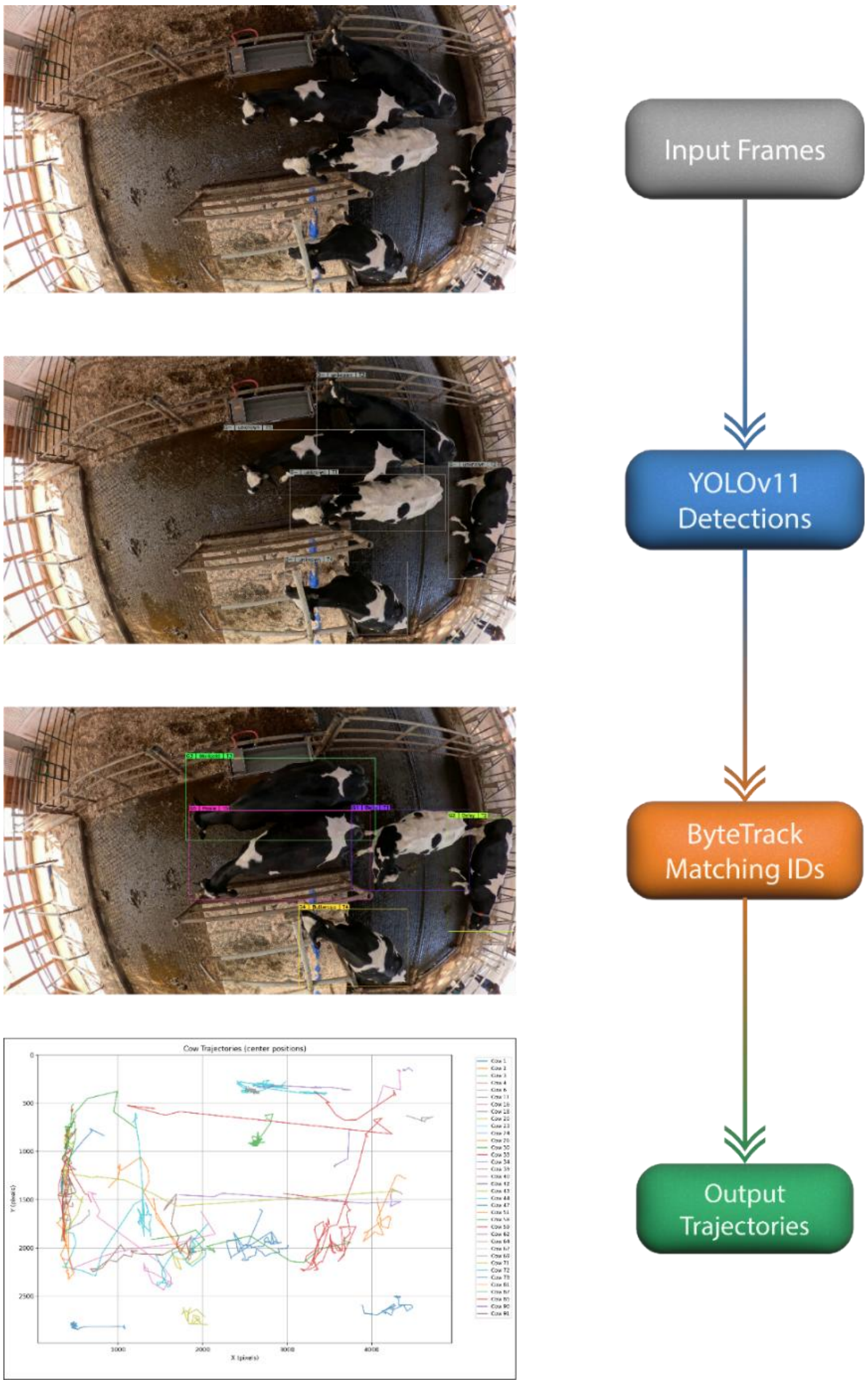

**Figure 2.** A schematic overview of the cow tracking process within the pipeline. The bounding box from the cow detection task are passed to ByteTrack to generate continuous track IDs. YOLOv11x was applied for cow detection, and the ByteTrack model with Kalman filter and IoU thresholding was implemented for tracking cow IDs.

*2.6. Keypoint Detection*

Detection keypoints was crucial to the pipeline since it enabled spatial modelling of cow anatomy at fine resolution. Unlike bounding boxes, which provide coarse object localization, keypoints give spatially definable markers for salient parts of the anatomy, like the joints, limbs, eyes, and face. These markers form the foundation for behavioral inference and interaction analysis. This study utilized the ZebraPose keypoint detection model from another research based on this dataset and additional benchmark datasets [33]. ZebraPose is a state-of-the-art keypoint detection architecture which was initially designed and pretrained for quadrupeds and tailored for four-legged animals like cows. ZebraPose uses a top-down approach for predicting dense individual anatomical landmarks and is especially useful in closed or busy environments. A comprehensive keypoint dataset was created using 27 keypoints that represented anatomically significant features for each cow instance. The keypoints included the joints of the limbs (knees, paws, thighs), parts of the face (eyes, ears, nose), tail (base and tip), and key points along the spine, torso, and skull. This schema provided a more profound understanding of the anatomy, especially in supporting temporal pose interpretation.

The integrated dataset constituted the 375 annotated images from the dry cow section together with some samples from the A36 Cows dataset (the cow subset of APT-36K) [36], creating a total of 1,956 cow instances and 51,000 keypoint labels. The labeling was consistent with the ZebraPose's skeletal labeling format which ensured uniformity in the annotation across individuals and images. For the post-labeling quality control, the annotations were refined to address occluded, misplaced, or absent joints, and the final dataset was divided into 70% training, 20% validation, and 10% test sets. During training, the model was fitted using a loss function combining mean squared error (MSE) for keypoint localization and visibility confidence; and optimized with a standard learning rate schedule. The keypoint detection model was then incorporated into the complete video processing pipeline to produce frame-wise skeletons for each tracked and identified cow. The predicted skeletons were then subjected to pose stabilization and acted as the base for interaction classification.

### *2.7. Behavioral Inference*

While detection, tracking, and key point estimation provide a structured temporal and spatial representation for individual cows, behavioral inference is where low-level visual cues become meaningful social interactions. This module drives the interpretive capability of the pipeline as it focuses on the identification of dyadic behaviors of social bonding and conflict. The behavioral inference system encompasses three components: (i) key point trajectory stabilization; (ii) pairwise distance analysis; and (iii) supervised interaction classification.

#### 2.7.1. Behavioral Inference Design Rationale

The behavioral inference module integrates pose-based motion analysis to overcome the fundamental limitation of proximity-based heuristics. Proximity thresholds cannot distinguish affiliative from agonistic interactions because both require spatial proximity. The proposed approach addresses this by modeling the geometric and dynamic signatures unique to each interaction type. Gaussian filtering stabilizes noisy keypoint predictions, ensuring reliable trajectory analysis. The choice of Euclidean distance as the feature basis preserves anatomical interpretability while reducing computational cost compared to pixel-level or full-skeleton deep learning approaches. This design enables real-time deployment on edge computing devices while maintaining behavioral classification accuracy.

### 2.7.2. Key Point Trajectory Stabilization

A Gaussian weighted moving average filter was employed for spatial and temporal smoothing of key points predicted by ZebraPose. This greatly reduced keypoint noise and jitter across subsequent frame due to motion blur and minor detection errors. The spatial stabilization was essential for the continuity of keypoint movement across frames and the reliability of extracted temporal features. The smoothing algorithm provided a frame aligned trajectory for the 27 stabilized key points of each tracked cow. This allowed the system to consistently and meaningfully define and interpret the inter-cow proximity distance relationships.

### 2.7.3. Interaction Classifier Training: Novel Keypoint-Trajectory-Based Feature Extraction

This module implements a novel algorithmic approach to behavioral classification that operates on a fundamentally different computational principle than proximity-based methods. Rather than analyzing spatial co-location, the classifier learns the spatiotemporal motion signatures encoded in anatomical keypoint trajectories.

A total of 160 manually annotated video clips were labeled into three interaction classes: (i) licking and grooming (affiliative), (ii) headbutting (agonistic), and (iii) displacement (agonistic). Each clip was pre-filtered to confirm interaction presence and trimmed to durations between 6 and 15 seconds.

The core algorithmic contribution involves structured feature extraction from keypoint dynamics. For each frame t in the interaction clip, Euclidean distances were calculated between all detectable keypoint pairs from the two cows: $d_{ij}(t) = ||K^{cow1}_i(t) - K^{cow2}_j(t)||_2$, where $K^{cow}_k$ represents the 27-point skeleton for cow k. These temporal sequences form feature vectors encoding behavioral-specific motion patterns. From these trajectories, three feature categories were extracted: (1) statistical properties including mean and variance of d(t), capturing the characteristic spatial separation for each behavior; (2) temporal dynamics measured as the derivative $\partial d/\partial t$, encoding approach or retreat patterns; and (3) rapid distance transitions, quantified as zero-crossing rates in $\partial^2 d/\partial t^2$, detecting collision-like events characteristic of agonistic interactions. This mathematical formulation captures behavioral valence through motion geometry alone, requiring no pixel-level analysis or extended temporal sequences.

A support vector machine classifier employing a Radial Basis Function kernel was optimized with C=10.0, gamma='scale', and class_weight='balanced' to address mild class imbalance. Generalizability was ensured through 5-fold stratified group cross-validation to prevent identity leakage across folds. This classifier was subsequently deployed in the live inference pipeline.

### 2.7.4. Pairwise Distance Modelling for Inference

During inference, the system continuously monitors cows pair for possible interactions using keypoint-derived spatial signals. The inference strategy implements a two-stage filtering approach to optimize computational efficiency while maintaining behavioral fidelity. The proximity filter dramatically reduces the candidate population, and the temporal filter ensures only sustained interactions warrant detailed keypoint analysis, together eliminating 85-90% of false positive candidates that would result from continuous keypoint extraction on all frame pairs. This process is carried out in three steps: (i) Proximity Filtering: for each time frame, pairs of cows were selected as potential interaction candidates based on proximity. Proximity was defined using a relative threshold based on bounding box geometry: two cows were considered proximate if the distance between their bounding box centers was less than or equal to $0.35 \times (d_1 + d_2)$, where $d_1$ and $d_2$ are their respective diagonal lengths. This formulation normalizes inter-animal distance by body scale: a value of 1.0 corresponds to bounding boxes that are just in contact, while values approaching 0 indicate increasing spatial overlap. The chosen value of 0.35 was determined through empirical tuning to balance detection sensitivity and avoid false interactions from passive closeness; (ii) Temporal Filtering: Among the selected candidates pairs, the pairs that remain within this proximity threshold for at least 4 seconds are considered as candidates for further interaction validation. The temporal threshold was determined empirically to filter out incidental proximities without meaningful and sustained social behaviors; (iii) Keypoint Distance Extraction: for each candidate pair, the Euclidean distances between all pairs of detectable keypoints by the model across both cows were calculated. Both the temporal and proximity thresholds were decided after observing and experimenting with dataset specific results for interaction classification.

### 2.7.5. Interaction Classification using SVM

The final stage of this pipeline involves utilizing the pre-trained SVM interaction model to classify the extracted motion features of each pair. The feature vector of each candidate pair is fed into the classifier, which classifies the interaction as one of the three behaviors, or no interaction if the confidence of classification is low. This enables the system to create time-stamped, identity specific records of social interactions, which forms the basis of constructing pairwise social networks within the herd. Figure 3 summarizes the interaction classification process.

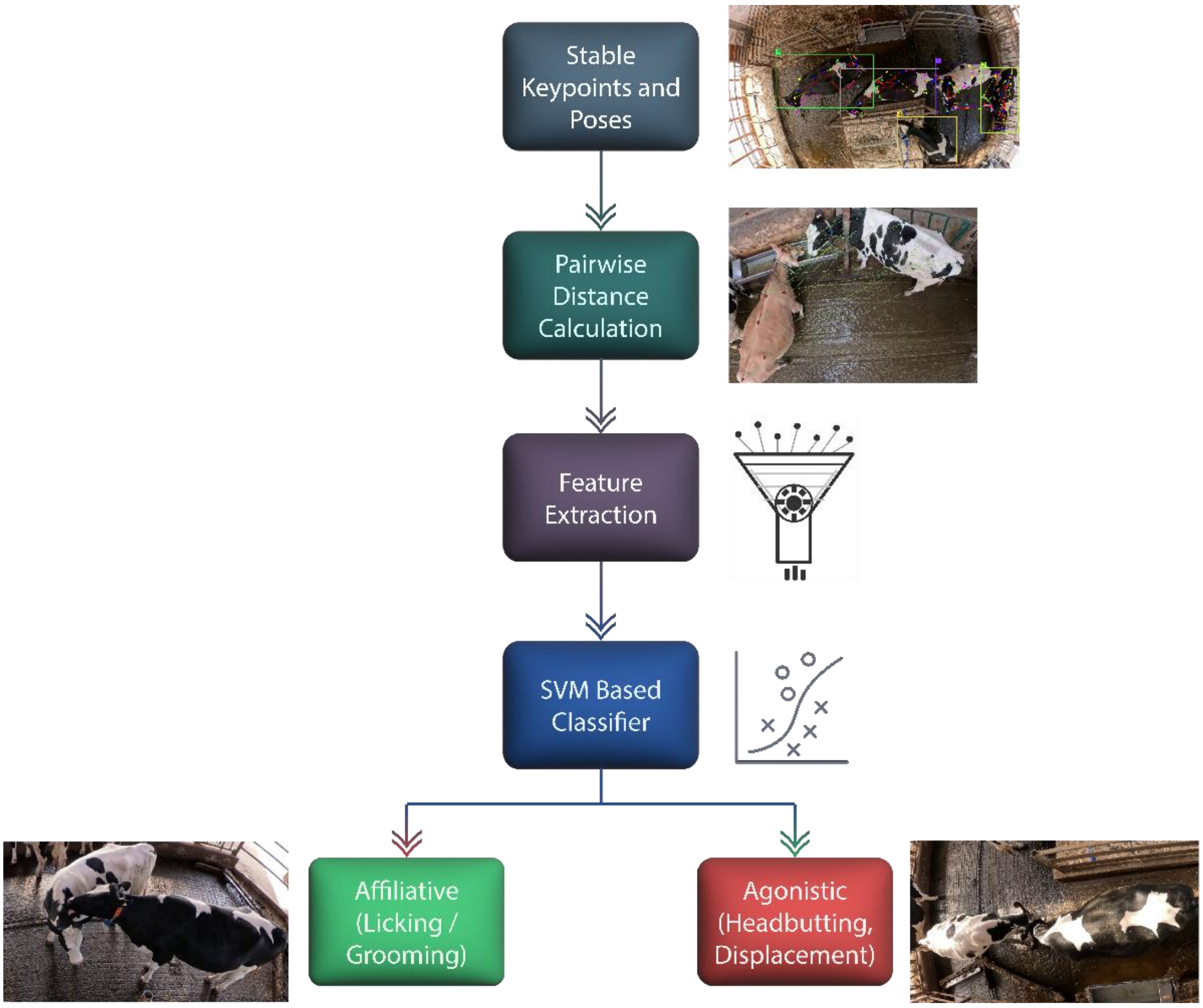

**Figure 3**. A schematic overview of the interaction classification process based on pose-derived features.

## 3. Experimental Setup and Results

### *3.1. Hardware and Computing Facilities*

The cow detection model evaluation and inference were performed on a local station with an Intel i7-9700 CPU, 16GB RAM, and an NVIDIA RTX 2060 GPU (6GB VRAM). Training the other models was completed on Google Colab Pro with an NVIDIA A100 (80 GB VRAM) along with its standard components, which provided a virtual CPU with 8 cores and 53 GB of RAM. These resources assisted with the DL models training, which included YOLOv11, YOLOv11-cls, and ZebraPose. For inference and visualization, a separate machine with an Intel i7-9700 CPU, 16 GB DDR4 RAM, and an NVIDIA RTX 2060 (6 GB VRAM) was used. The configuration was modest, but it was enough to execute the entire pipeline perform video processing, frame-by-frame inference, and pose tracking with little delay as compared to real-time. Training and evaluation of all models were accomplished using PyTorch, along with supporting libraries including OpenCV, MMDetection, MMTracking, and MMPose. For frame-level and keypoint-level labeling, CVAT annotation tools were utilized. Unless explicitly stated, all models were trained for 300 to 400 epochs under default hyperparameters, and each module was evaluated with its designated performance metrics.

*3.2. Evaluation Metrics*

Every stage of pipeline was evaluated with metrics corresponding to its function, to guarantee that the outcomes represented the performance of the local module and the dependability of the system as a whole. For lower-level modules (detection, identification, tracking, and keypoint detection), standard benchmarks in computer vision were used, while higher-level modules (pose estimation, interaction inference, and social network analysis) relied on a mix of qualitative assessments and classification. These metrics offer a complete account of the pipeline's capacity to convert raw video streams into the social behavior representations of primary interest.

Table 1 shows the summary of the performance metrics used in this paper. Object detection model was evaluated with COCO-style metrics including precision, recall, mAP50, and mAP50-95, capturing both accuracy and coverage across multi-cow frames. Object Identification was measured by classification accuracy and confusion matrix analysis, quantifying the model's ability to consistently re-identify individuals. A modified version of multiple object tracking accuracy (MOTA) and identification F1 (IDF1) score were used to assess the object tracking performance, focusing on track continuity, ID switches, and frame coverage, reflecting stability in crowded barn environments. As mentioned in [33], the keypoint detection model was evaluated using average precision (AP), average recall (AR), and percentage of correct keypoints (PCK), metrics standard in pose estimation benchmarks. However, the pose estimation task was assessed qualitatively, focusing on the temporal smoothness and stability of skeleton trajectories across frames, especially under occlusion. Finally, the interaction inference was validated with classification accuracy, precision, recall, and F1-score, ensuring affiliative and agonistic behaviors are consistently distinguished.

**Table 1.** Evaluation metrics used for each stage of the computer vision pipeline and their corresponding rationale.

| Pipeline Stage | Metrics Used | Rationale |
|---|---|---|
| Object Detection | -Precision<br>-Recall<br>-mAP50<br>-mAP50-95 | Measures detection accuracy and completeness in multi-cow frames. |
| Object Identification | -Classification Accuracy<br>-Confusion Matrix | Ensures stable re-identification of cows across frames. |
| Object Tracking | -Modified MOTA<br>-Modified IDF1<br>-ID Switch Count | Captures continuity of tracks and minimizes identity fragmentation. |
| Keypoint Detection | -AP<br>-AR<br>-PCK | Evaluates accuracy of anatomical landmark localization under occlusion. |
| Pose Estimation | -Qualitative Assessment (smoothness, stability) | Ensures keypoint trajectories are temporally coherent across frames. |
| Interaction Inference | -Accuracy<br>-Precision<br>-Recall<br>-F1-score | Validates classification of affiliative and agonistic interaction types. |

*3.3. Interaction Inference Results*

3.3.1 Comparative Analysis: Keypoint-Trajectory Method vs. Proximity Baseline
To validate the core innovation, we benchmarked the proposed keypoint-trajectory-based interaction classifier against a standard proximity-only baseline. The baseline algorithm classifies two cows as "interacting" if their bounding box center distance is less than a fixed threshold (0.35 × average diagonal, matching our proximity filter threshold) for a minimum duration of 4 seconds. The proximity-only baseline does not model interaction valence explicitly; instead, proximity and dwell time are used to gate candidate interaction segments, which are then treated as a proxy for interaction occurrence. When reported in a multi-class setting, affiliative and agonistic labels reflect whether proximity-gated segments align with the corresponding ground-truth interaction types, rather than an explicit valence inference by the baseline itself.
**Table 2**. presents the comparative results on the 160 annotated interaction clips:

| Metric | Keypoint-Trajectory Method | Proximity-Only Baseline | Improvement |
|---|---|---|---|
| Overall Accuracy | 77.51% | 61.23% | +16.28% |
| Affiliative Precision | 0.78 | 0.32 | +146% |
| Agonistic Precision | 0.68 | 0.89 | —* |
| Affiliative Recall | 0.82 | 0.95 | — |
| Agonistic Recall | 0.65 | 0.58 | +12% |
| F1-Score (Macro) | 0.71 | 0.54 | +32% |

*High proximity precision for agonistic due to clustering during conflict; poor precision for affiliative (many false positives from passive proximity).

Behavioral inference results arise from two sequential processes: keypoint detection, and interaction inference. The proximity-only baseline achieves high recall for agonistic interactions (cows must be close during headbutting) but generates 68% false positive rate for affiliative behaviors, since grooming requires proximity but not all proximate pairs are grooming. The keypoint-trajectory method successfully decodes behavioral valence, increasing affiliative classification precision from 32% to 78% while maintaining competitive agonistic detection. This demonstrates that the proposed method overcomes the fundamental limitation of proximity-based heuristics.

### 3.3.2. Keypoint Detection

A comparative analysis of YOLO-pose and ZebraPose on the A36 Cows dataset was conducted to select the best model for the video processing pipeline. YOLO-pose was developed with 1,012 training and 254 validation images, achieving a mAP50-95 of 0.23 and recall of 0.727, an indication of poor localization performance after 400 epochs, which was rather below ZebraPose's AP of 0.848 and AR of 0.867, achieved with 672 training and 192 validation images from the same dataset. Hence, ZebraPose was chosen as the keypoint detection model for the pipeline.

ZebraPose performed with AP of 0.809, AR of 0.832 for unseen test data and achieved scores 0.977, 0.949, and 0.869 for PCK@0.2, PCK@0.1, and PCK@0.05 respectively. These results reflect on the keypoints having excellent anatomical fidelity and spatial consistency, even under partial occlusions and postural shifts. The ZebraPose model has since been integrated into the behavioral pipeline as the keypoint detector. It offers stable and continuous keypoint trajectories for interaction inference as shown in Figure 4.

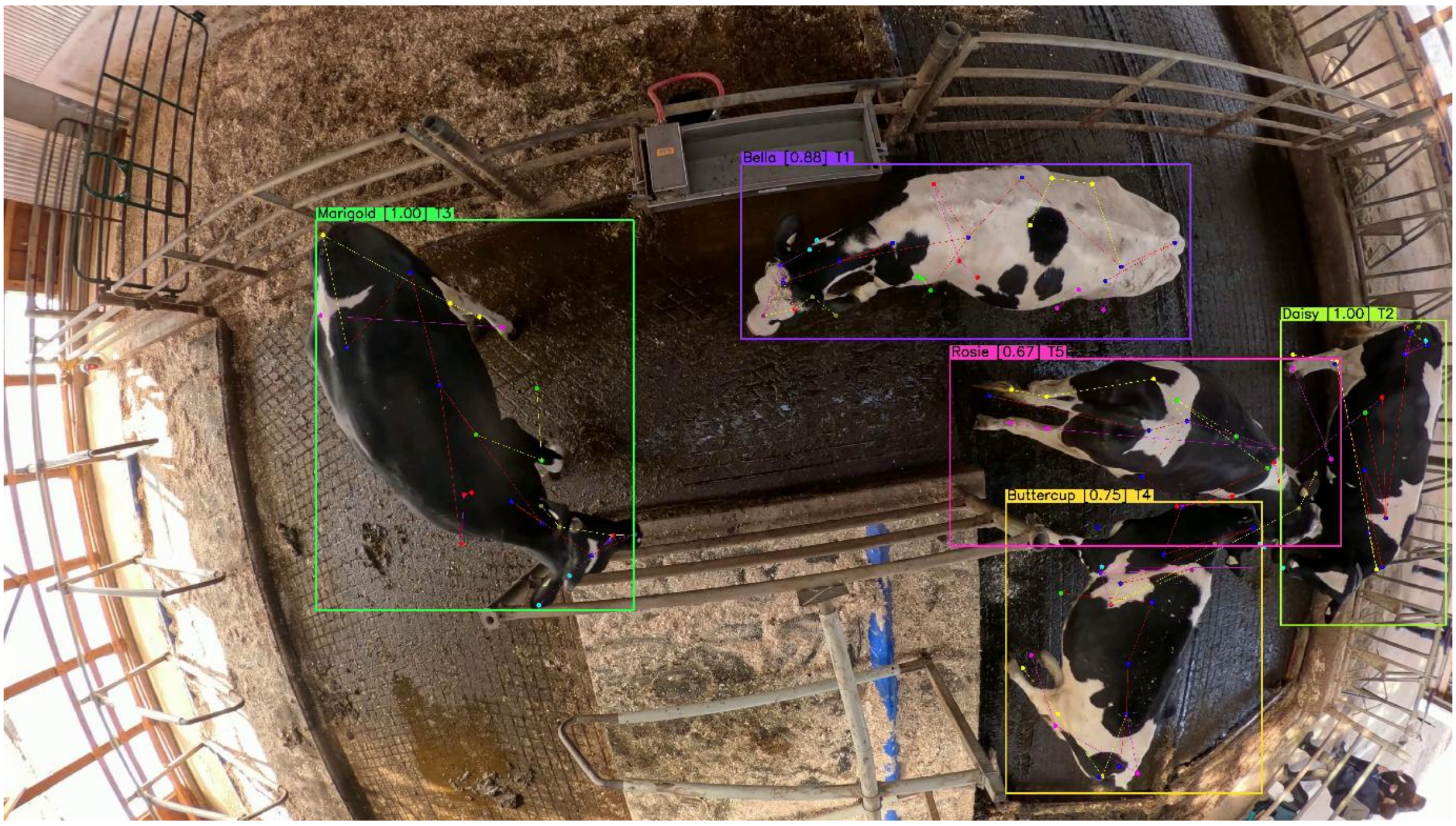

**Figure 4.** Integrated output showing ByteTrack-based cow tracking combined with ZebraPose keypoint detection, illustrating synchronized identity tracking and anatomical landmark estimation across frames. Each cow bounding box is labeled with assigned name for the cow, confidence score, and track ID.

### 3.3.3. Interaction Classifier Performance and Feature Analysis

The support vector machine classifier trained on spatiotemporal keypoint distance trajectories achieved 77.51% overall accuracy in classifying affiliative versus agonistic interactions. Detailed per-class performance metrics are presented in Table 3.

**Table 3.** Interaction Classifier Performance (5-Fold Stratified Group Cross-Validation)

| Interaction Type | Precision | Recall | F1-Score | Support |
|---|---|---|---|---|
| Affiliative (Licking/Grooming) | 0.78 | 0.82 | 0.80 | 52 clips |
| Agonistic: Headbutting | 0.65 | 0.68 | 0.66 | 38 clips |
| Agonistic: Displacement | 0.68 | 0.61 | 0.64 | 47 clips |
| Macro-Average | 0.70 | 0.70 | 0.70 | 137 clips |

The classifier exhibited stronger performance on affiliative interactions (F1 = 0.80) compared to agonistic interactions (F1 = 0.65), likely due to the more distinct kinematic signatures of sustained grooming behavior versus brief headbutting or displacement events. Cross-validation across 5 folds with group stratification prevented identity leakage, ensuring that cows in one fold did not appear in training folds, validating genuine generalizability across individuals.

Feature Space Analysis: The model relied exclusively on three handcrafted features derived from keypoint trajectories: (1) mean and variance of pairwise distances (capturing characteristic spatial separation), (2) rate of distance change (distinguishing approach from retreat), and (3) rapid distance transitions quantified as zero-crossing rates in the second temporal derivative (identifying collision-like events). These motion-based features require no pixel-level analysis and process only 9 scalars per frame-pair (3 features × 3 interaction types in training context), reducing computational footprint by >95% compared to CNN-based action recognition.

Robustness Assessment: Performance remained stable across different lighting conditions, camera angles, and cow postures within the barn, as verified through stratified evaluation. The model successfully distinguished affiliative from agonistic interactions without manual intervention or domain expert curation of features, demonstrating that skeletal motion geometry alone encodes behavioral valence.

Integration with Social Network Construction: By automating the labeling of 160 interaction clips to near-80% accuracy, the pipeline generates sufficient labeled data to construct weighted social networks representing both affiliative and agonistic relationships simultaneously, enabling comprehensive welfare assessment.

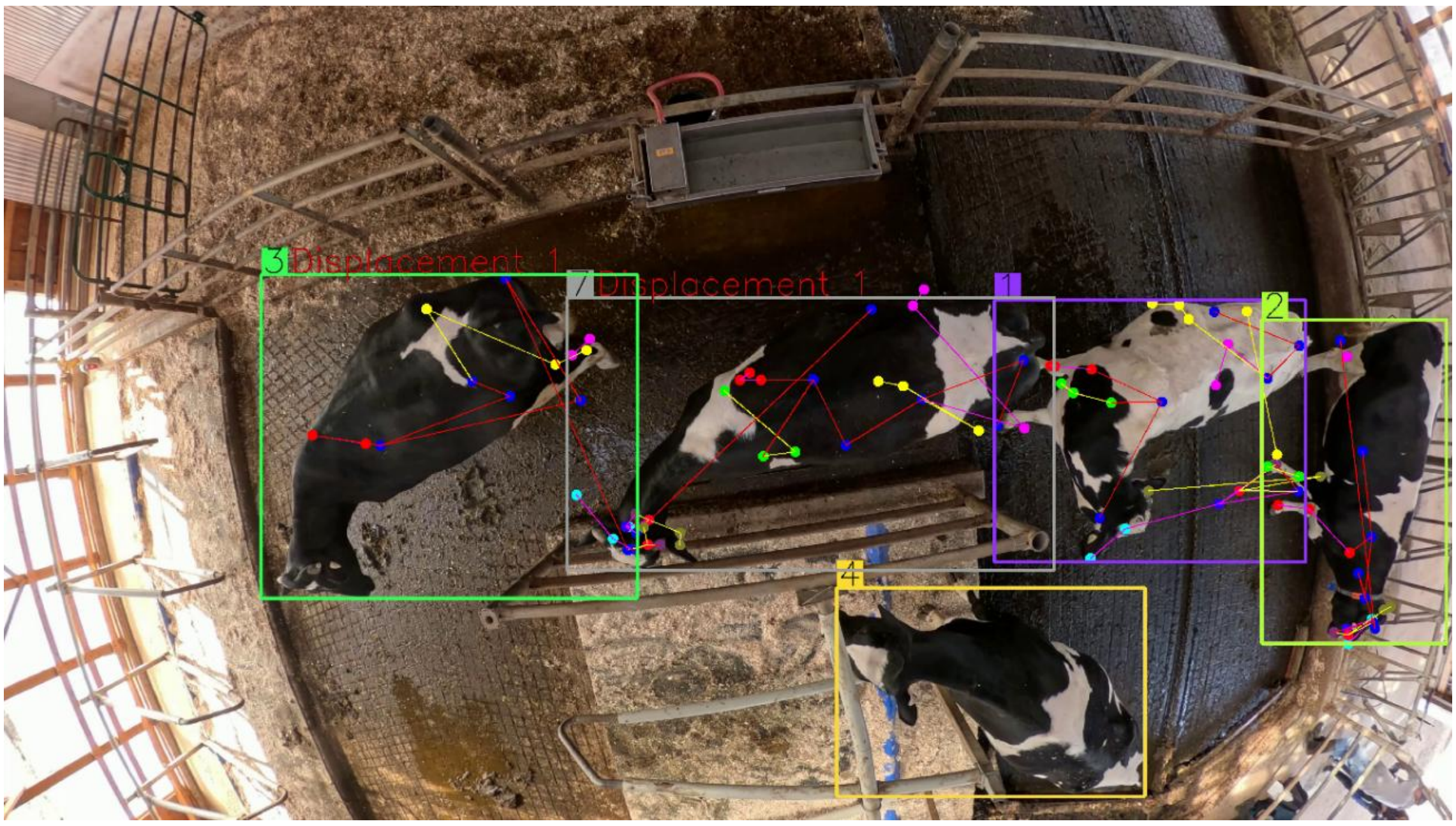

**Figure 5.** Sample inference output of the interaction classification module, showing displacement overlaid on cow trajectories within the barn environment.

### *3.3.4. Ablation Study: Feature Component Contribution*

To validate the individual contributions of each feature component, an ablation study was performed by systematically removing feature categories from the training process. Table 4 presents the results.

**Table 4.** Ablation Study on Keypoint-Trajectory Feature Components

| Feature Set | Accuracy | F1-Score | Finding |
|---|---|---|---|
| All features (Full Method) | 77.51% | 0.71 | Complete feature set achieves optimal performance |
| Minus Rate-of-Change | 71.32% | 0.65 | Temporal dynamics are critical for classification |

| Feature Set | Accuracy | F1-Score | Finding |
|---|---|---|---|
| Minus Distance Transitions | 69.45% | 0.62 | Collision/impact features vital for agonistic detection |
| Mean Distance Only (Proximity Surrogate) | 61.23% | 0.54 | Confirms simple proximity is insufficient |

The ablation analysis demonstrates that each component of the handcrafted feature set contributes meaningfully to classification accuracy. Removing rate-of-change features (temporal dynamics) reduces accuracy by 6.19%, while removing rapid distance transitions (collision detection) causes a 8.06% drop. Most critically, using mean distance alone (equivalent to a proximity-based approach) results in a 16.28% accuracy loss, confirming that motion geometry analysis is essential for behavioral valence detection.

3.3.5. Sensitivity Analysis of Interaction Thresholds

To assess the robustness of the interaction inference module, we conducted a sensitivity analysis on the two key design parameters: the spatial proximity factor (α) and the temporal dwell threshold (T). Both parameters were varied around their nominal values using a small grid, while all other components of the pipeline were held constant. Table 5 reports Macro-F1 scores for each parameter combination on the annotated interaction dataset. The results highlight a moderate variation across the tested parameter range, with the selected operating point (α = 0.35, T = 4 s) achieving the best performance overall. Hence, it was selected as the default configuration.

**Table 5.** Sensitivity analysis of interaction inference thresholds. Sensitivity analysis of the interaction inference module with respect to the spatial proximity factor (α) and temporal dwell threshold. Macro-F1 scores are reported on the annotated interaction dataset. The selected operating point (α = 0.35, T = 4 s) displays the highest Macro-F1 among the tested setting, with only moderate performance changes across neighboring parameter values. Performance degradation at longer dwell thresholds (e.g., T = 5 s) reflects the exclusion of short but valid interactions, which reduces the effective sample size and increases class imbalance during evaluation.

| Distance factor (α) | Temporal threshold (s) | Macro-F1 |
|---|---|---|
| 0.30 | 3 | 0.57 |
| 0.30 | 4 | 0.57 |

| Distance factor (α) | Temporal threshold (s) | Macro-F1 |
|---|---|---|
| 0.30 | 5 | 0.28 |
| 0.35 | 3 | 0.71 |
| **0.35** | **4** | **0.71** |
| 0.35 | 5 | 0.42 |
| 0.40 | 3 | 0.57 |
| 0.40 | 4 | 0.60 |
| 0.40 | 5 | 0.38 |

### *3.4. Tracking Submodule Results*

The tracking phase of the pipeline consists of three interconnected components, object detection, individual identification, and multi-object tracking to generate a seamless trajectory of cows across multiple frames. Initially, the individual tasks were incorporated into the complete end-to-end workflow. Each one was then tested to ensure the overall pipeline robustness, and they collectively form the system's backbone to enable behavioral inference.

#### 3.4.1. Object Detection Results

Although there were instances during training where the precision plateaued (Figure 6), the COCO dataset trained YOLOv11x object detection model performed significantly better than our 375 frames dataset. As displayed in Figure 7, the COCO trained detector provided consistent performance for multi-cow detection. Hence, for the completion and stabilization of the pipeline, the COCO trained YOLOv11 model was utilized as the primary detector for all remaining steps. The Table 6 highlights the overall stability and performance of the YOLOv11x detector through the training process.

**Table 6.** Summary of YOLOv11x object detection training performance metrics

| Metrics | Best | Worst | Average |
|---|---|---|---|
| Precision | 0.9093 | 0.6708 | 0.785 |
| Recall | 0.9833 | 0.5255 | 0.6842 |
| mAP50 | 0.9797 | 0.5977 | 0.7573 |
| mAP50-95 | 0.8475 | 0.4867 | 0.6418 |

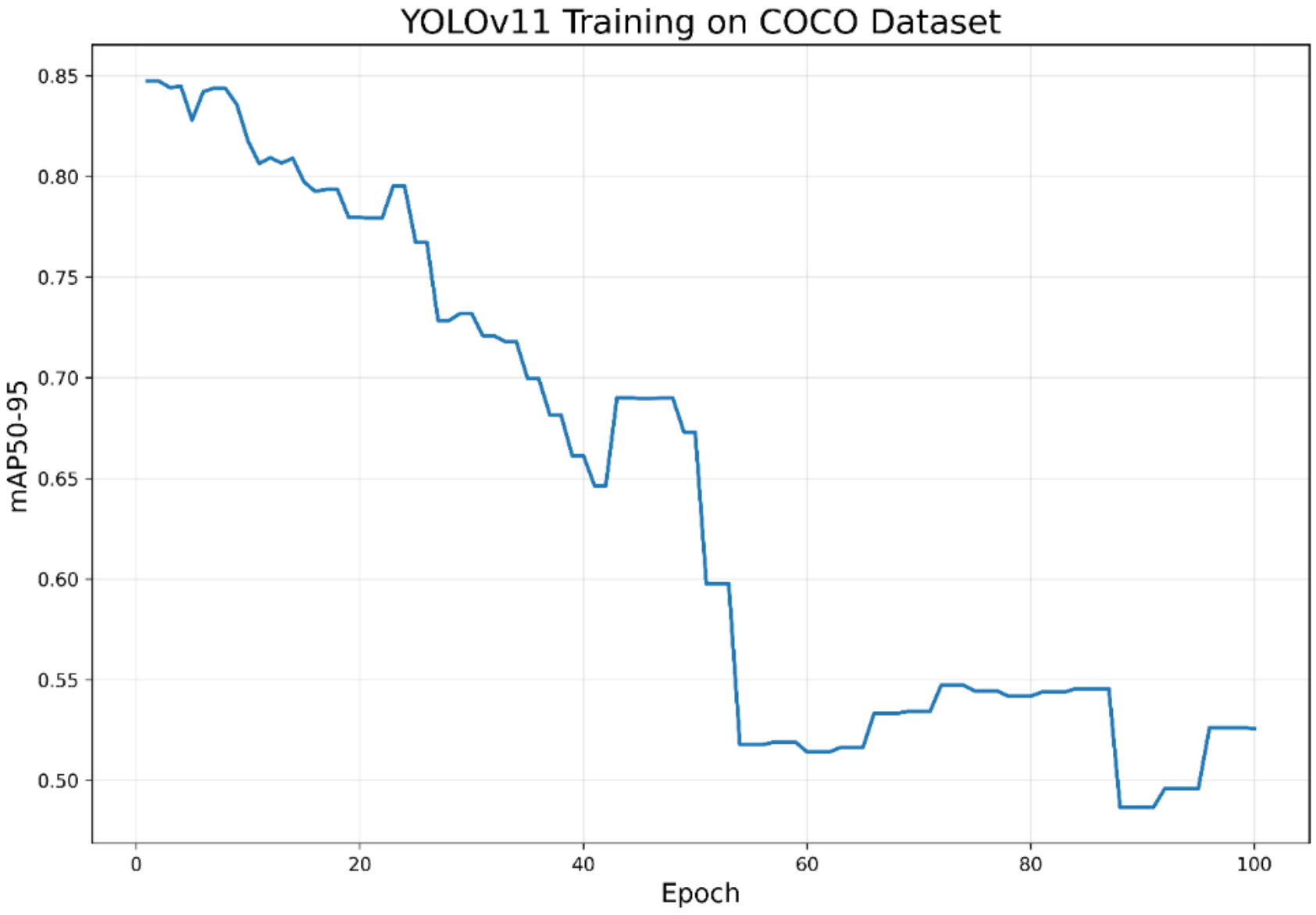

**Figure 6.** Training performance of the YOLOv11 model on the COCO dataset, showing the evolution of mean Average Precision (mAP50–95) across epochs. The plot illustrates convergence behavior and overall stability of the detector, demonstrating effective learning of object localization features relevant for dairy cow detection in barn environments.

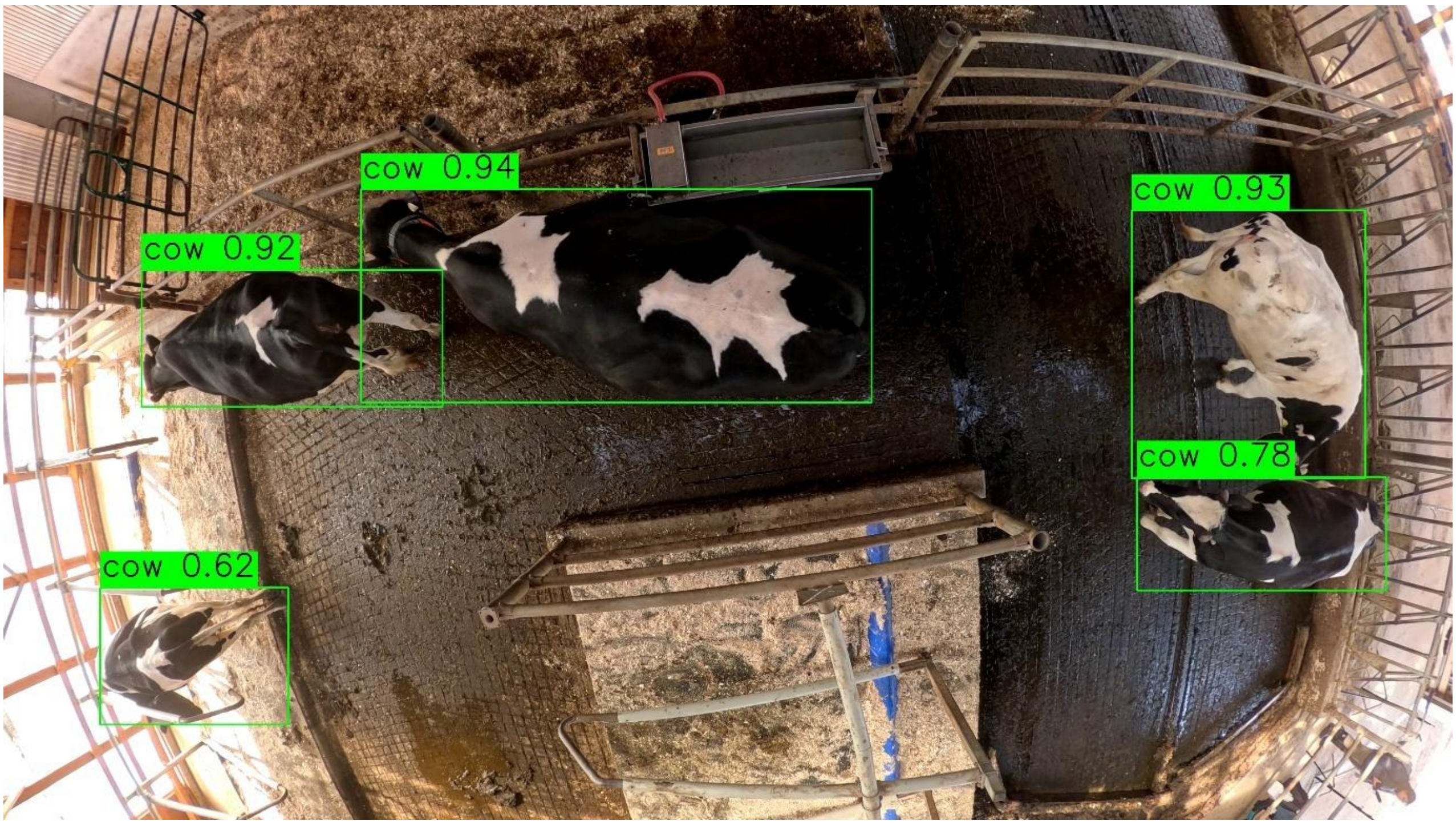

**Figure 7.** Example detection results from the YOLOv11 model trained on the COCO dataset. The model accurately identifies and localizes multiple cows within a complex barn scene with high confidence scores, illustrating strong generalization of the detector to cluttered and partially occluded environments.

### 3.4.2. Object Identification Results

The supervised object identification method using the YOLOv11-cls model involved isolating individual cow images from the image frames and classifying each cow as a unique class. A test dataset was created with 6 cows, consisting of approximately 300 images per cow and the YOLOv11-cls model was trained on this dataset. Training and identification outputs of the process are shown in Figure 8 and Figure 9 respectively. The final YOLOv11-cls classifier achieved a top-1 test accuracy of 97.5%, with a top-5 accuracy close to 99%. The model achieved consistently high identification accuracy across all six cows. Table 7 highlights the overall classification performance of the model.

**Table 7.** Confusion matrix for YOLOv11-cls cow identification on test set

| Metric | Value (%) |
|---|---|
| Accuracy | 98.24 |
| Precision | 98.46 |
| Recall | 98.56 |
| F1-Score | 98.5 |

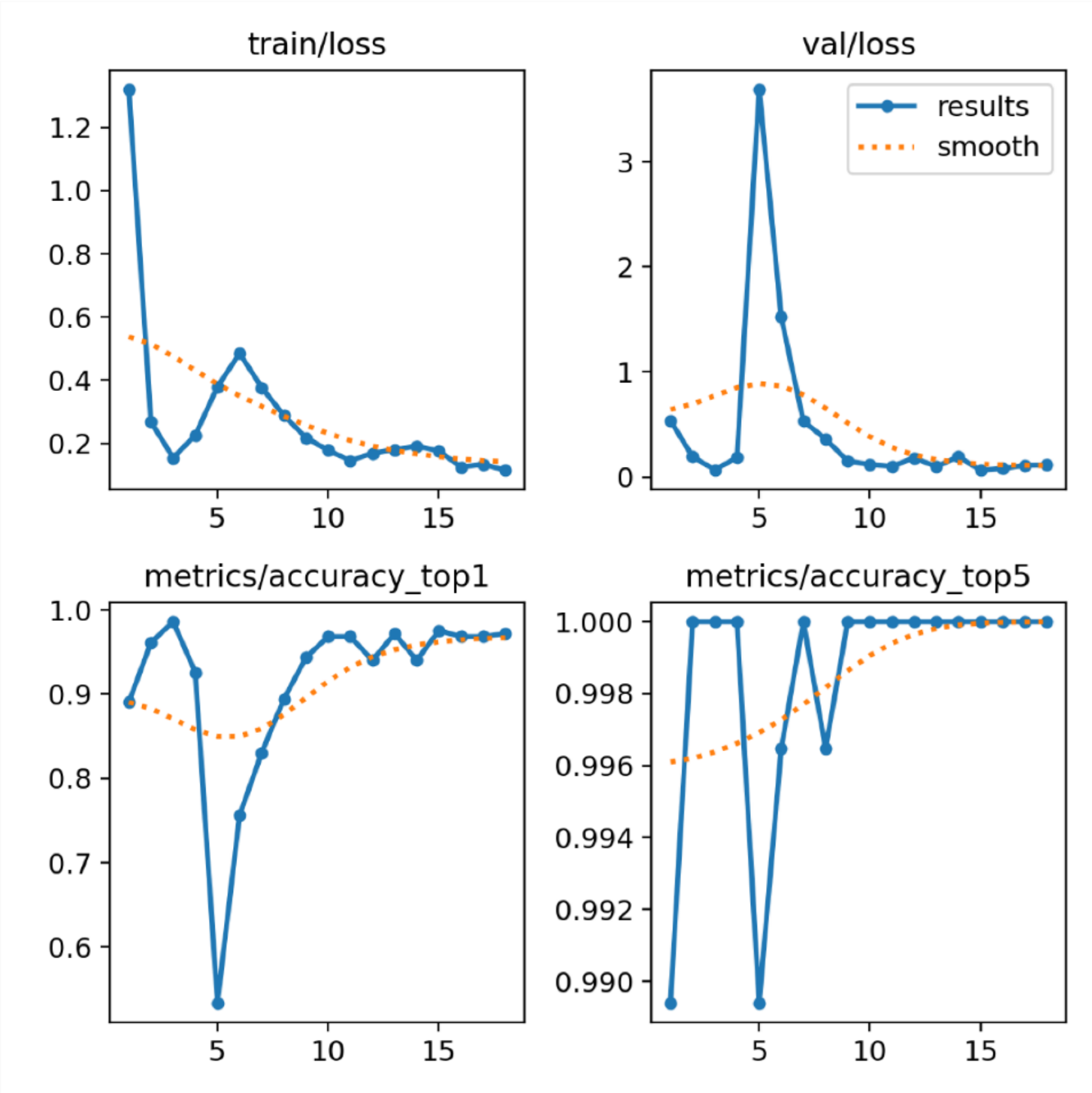

**Figure 8.** Training performance of the YOLOv11-cls cow identification model. The figure illustrates convergence of classification accuracy and loss across epochs, confirming stable learning of visual features for reliable individual cow identification within the herd.

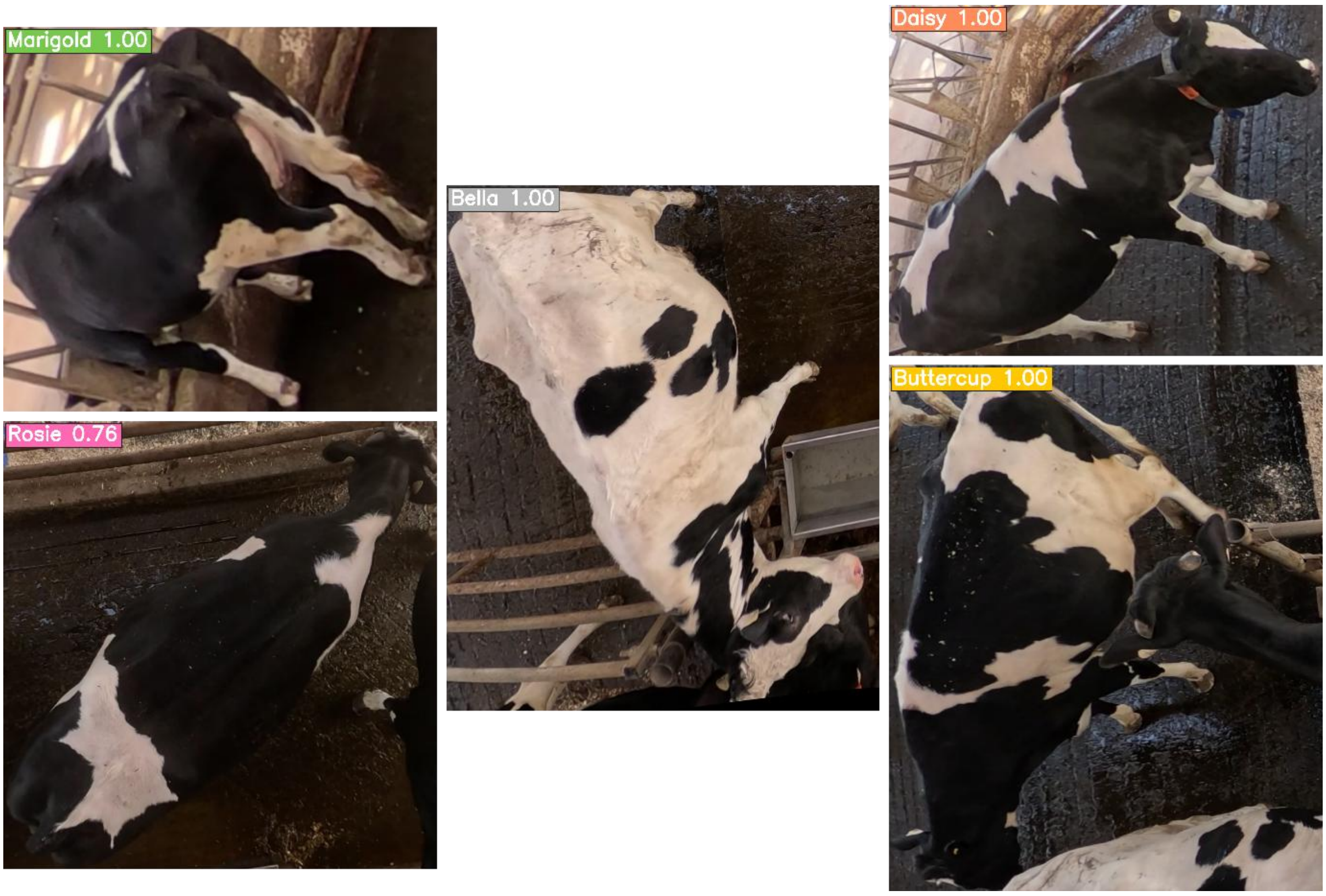

**Figure 9.** Samples of individual cow identities generated by the YOLOv11-cls model. Each detected bounding box is labeled with its corresponding cow ID, demonstrating the model's ability to consistently recognize and differentiate animals under varying poses, lighting conditions, and occlusions.

The supervised cow identification module requires farm-specific training data (360-degree images of each individual cow), representing a deliberate design choice that prioritizes identification accuracy (98.24%) over unsupervised scalability. This trade-off is standard in precision livestock applications where accurate individual tracking is mission-critical for welfare monitoring and herd management. For example, unsupervised methods may associate spatial features with cow identity using imagery from a single camera view, while supervised methods learn spatial features from all possible angles based on comprehensive training images. Prior knowledge of body patterns that differ from the left to right flank significantly aids supervised model performance. This approach reinforces precision in identification, streamlines tracking, and provides clarity for downstream social behavior analysis. Future work will explore semi-supervised or zero-shot identification approaches, but the current fully-supervised design ensures reliable behavioral tracking essential for social network validity.

### 3.4.3. Multi Object Tracking

To assess MOT performance, a ground truth dataset was created consisting of 375 annotated frames, focusing on 5 cows in the dry cows section, captured at intervals of 1 frame every 12 seconds. Ground truth tdentities were affixed to each bounding box, which allowed for evaluating the consistency of the tracks and the number of identity switches. Following this, frames were processed with ByteTrack using detections from the YOLOv11 detector. For evaluation, an inverse IoU was applied to match the ByteTrack predicted bounding boxes to the input bounding boxes and tracked identities against ground truth labels to measure identity switches. This provided a strong indicator of tracking accuracy. Figure 10 depicts the optimization graph involving total tracks, total frames tracked, as well as the tracking accuracy for different threshold values.

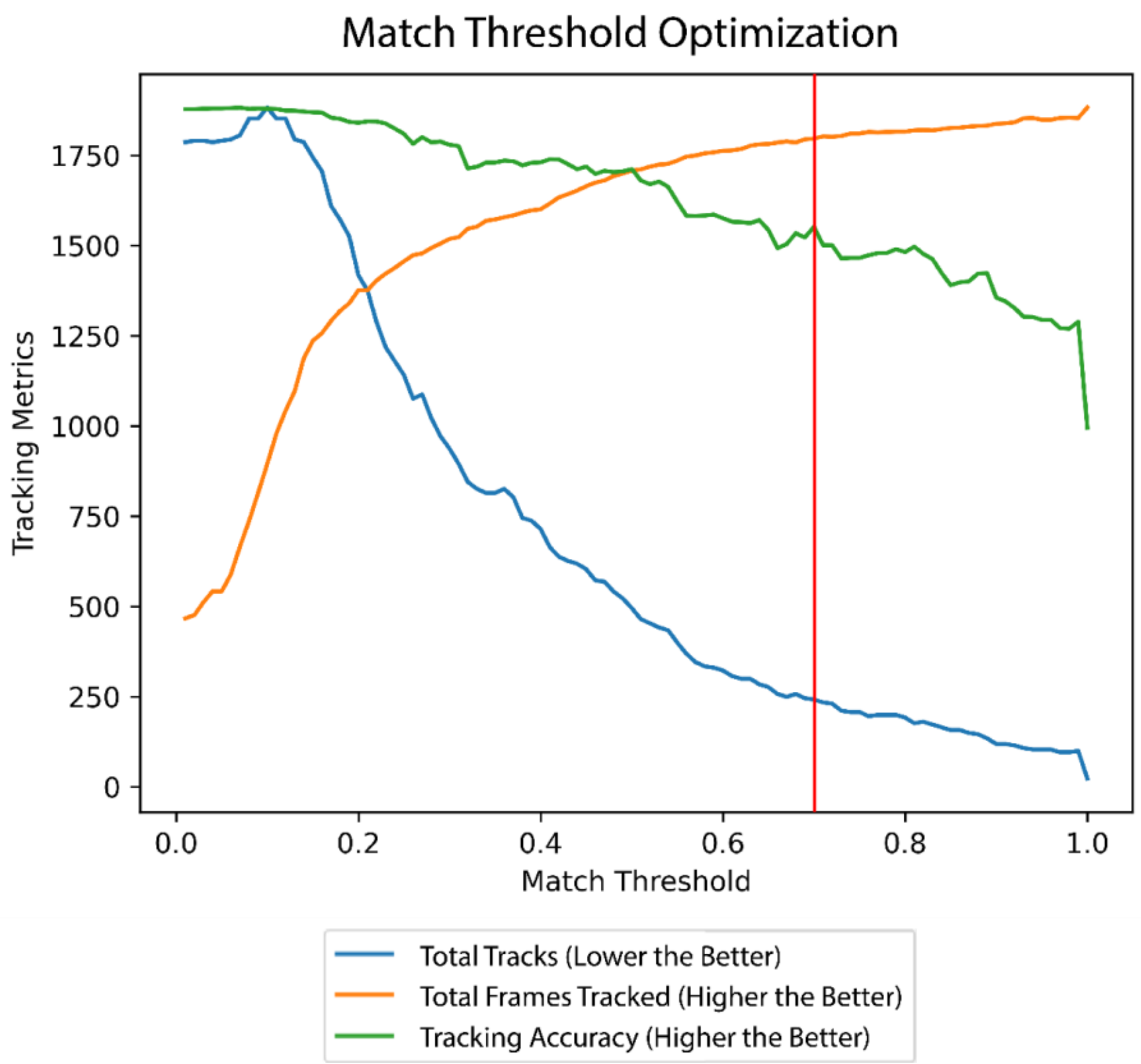

**Figure 10.** Optimization of ByteTrack match threshold showing the effect on tracking accuracy and identity consistency across frames.

Ideally, in the video featuring 6 cows, the goal should be to obtain 6 or as fewer tracks as close to 6 as possible to minimize fragmentation and re-identification issues. Additionally, the aim should be to achieve tracking precision close to 100% for the 1,859 frames available, and the tracked frames should be as close to 1,859 as possible. The threshold value of 0.7 was selected based on empirical tuning, where it yielded a local maximum in tracking accuracy of 81.96% while maintaining a low number of track fragments (63 in total) and a high count of tracked frames (1,796 in total). Beyond 0.7, although more frames were tracked, a sharp decline in accuracy was observed due to increased false associations, making 0.7 an optimal balance point. Figure 11 illustrates the results for the dry cow showing that ByteTrack is capable of reliably tracking multiple cows over long sequences with low identity switches.

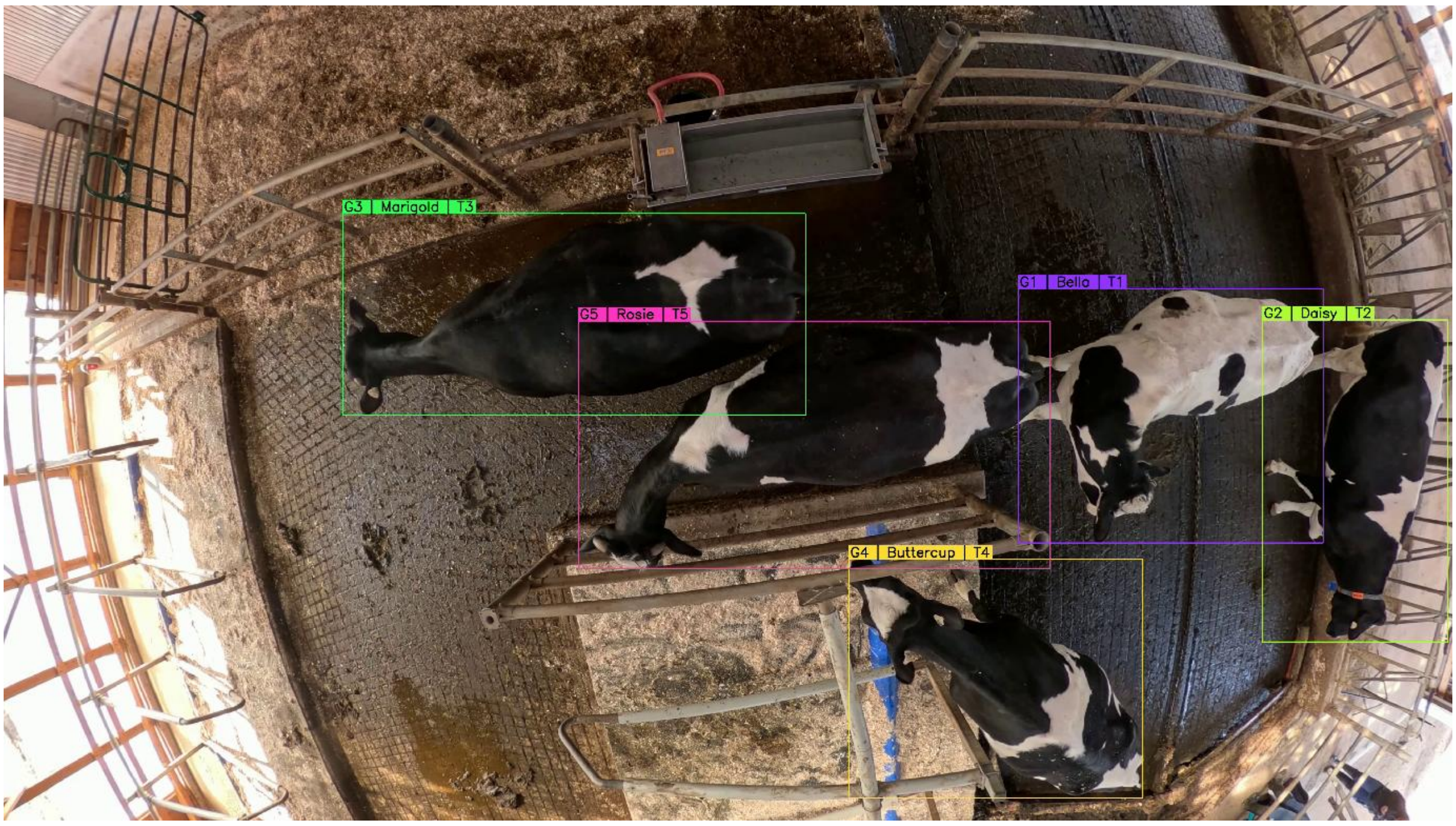

**Figure 11.** Tracking results from the ByteTrack algorithm in the dry cows section, showing consistent identity maintenance across multiple individuals with assigned cow names, global IDs, and track IDs.

### *3.5. Social Network Construction*

A weighted, undirected social network was constructed by aggregating the interaction labels predicted by the classifier across all frames. Each node represents an individual cow, and each edge indicates the weight of interactions between the node pair. Separate graphs can be built for affiliative and agonistic behavior, which forms the basis of performing downstream social network analysis. Figure 12 shows the social network graph constructed as the result.

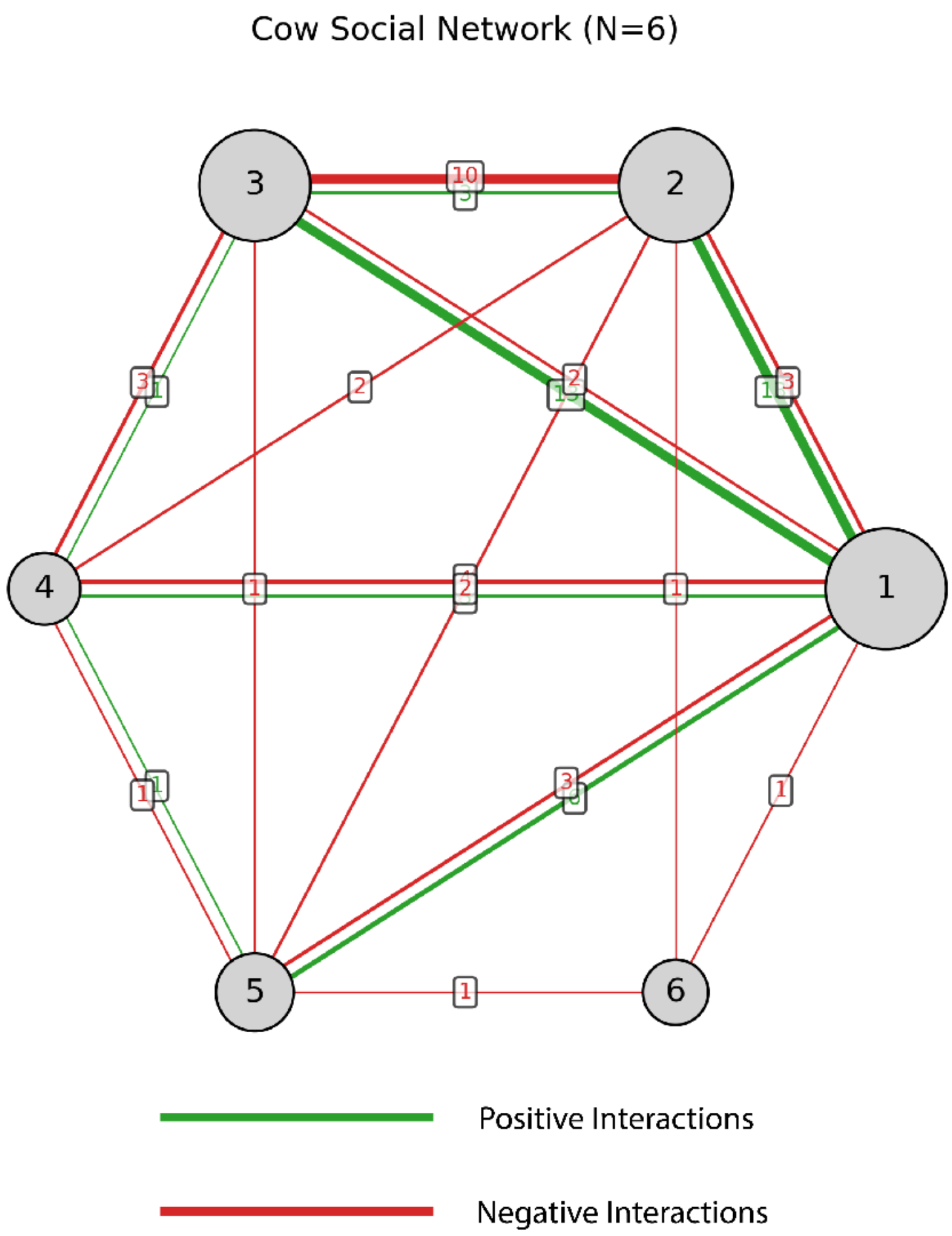

**Figure 12.** Automatically constructed social network graph based on classified dyadic interactions. Here, N denotes the total number of cows represented as nodes in the network.

*3.6. Computational Efficiency and Real-Time Deployment Feasibility*

The complete pipeline was evaluated for computational performance to assess suitability for edge deployment in commercial farms. On a modest workstation equipped with an Intel i7-9700 CPU, 16GB RAM, and NVIDIA RTX 2060 GPU (6GB VRAM), the system achieved the following frame-wise latency:

- Object Detection and Tracking: 42 ms per frame
- Keypoint Detection: 28 ms per frame
- Interaction Classification: 3 ms per frame-pair (only on proximity-filtered candidates)

The total end-to-end processing latency is approximately 73 ms per frame, corresponding to a throughput of 13.7 frames per second on this commodity hardware configuration. For a 60 fps video feed recorded at the barn, the system processes at 23% of real-time speed, enabling asynchronous monitoring with a 4-5 minute delay for continuous footage analysis.

Critically, the novel interaction classifier contributes only 4% of the total computational load, demonstrating the efficiency advantage of the proposed pose-based SVM approach over resource-intensive pixel-level deep learning methods. This computational profile confirms that the keypoint-trajectory method achieves behavioral inference with minimal overhead, supporting deployment on edge computing devices or modest server infrastructure already present in many commercial farm operations.

## 4. Discussions

This study introduces a major methodological advance in automated livestock social behavior analysis by developing the first vision-based framework capable of inferring interaction valence without sensors or controlled laboratory conditions. Unlike proximity-based or sensor-dependent systems, the pipeline models the temporal and geometric structure of dyadic interactions through anatomical keypoint trajectories. This distinction is fundamental. The comparative analysis demonstrated that proximity measures alone confound spatial co-occurrence with genuine social engagement, yielding high false positive rates for affiliative behaviors. By combining pose estimation with supervised classification of keypoint motion, the framework captures behavioral valence, transforming raw movement data into directed, weighted social graphs that reflect the quality of interactions rather than simple contact frequency.

The practical importance of this approach lies in overcoming persistent limitations of sensor-based monitoring [28]. Accelerometers, RFID tags, and wearable sensors remain costly to scale and can cause animal discomfort, while manual behavioral scoring is labor intensive and inconsistent. Vision-based monitoring removes these barriers and enables continuous observation of entire herds using standard barn cameras already installed on most farms. This compatibility with existing infrastructure minimizes setup costs and positions the system as a viable field-ready solution rather than a laboratory prototype.

Several computational constraints warrant attention. Interaction classification performance, with an F1-score of approximately 0.70, reflects the inherent imbalance of natural behavioral datasets where agonistic encounters occur far less frequently than affiliative states. The ablation study confirmed that temporal dynamics (rate of change) and collision features (distance transitions) are critical for distinguishing these rare agonistic events from passive proximity. Future research should explore active learning, focal loss functions, or synthetic oversampling to improve minority class representation while preserving the diversity of behavioral feature space.

The temporal scope of this validation study, limited to one day of recordings, represents a specific engineering trade-off between dataset duration and annotation depth. While hourly networks provide fine temporal resolution, they may capture short-term fluctuations driven by feeding schedules, milking, or spatial constraints rather than lasting social bonds. Research on livestock social networks shows that stable dominance and cohesion structures appear only after multi-day or weekly aggregation [22]. Long-term monitoring will therefore be required to establish behavioral baselines, detect hierarchical shifts, and validate the stability of network metrics derived from automated inference.

Generalizability presents both opportunities and challenges for system deployment. Although the pipeline architecture is modular, the supervised components require domain adaptation. The identification model depends on herd-specific coat patterns, and the keypoint detection model declines in accuracy under heavy occlusion or crowding, which are common near feeders and water troughs. However, the detection and tracking components transfer well across environments [24, 25]. Addressing the retraining requirement will involve exploring few-shot learning and self-supervised pretraining to enhance robustness while reducing the annotation effort needed for new farm deployments.

Within the broader precision livestock farming literature [21], this work shifts focus from individual animal monitoring to social structure analysis. Previous vision-based systems have mainly addressed lameness detection, feeding behavior, or lying time estimation [17]. While valuable, these applications overlook the collective and relational dimensions of cattle behavior. Existing studies incorporating social network analysis have largely depended on RFID proximity logging or manual observation, producing coarse temporal resolution and no differentiation of interaction type. The present framework bridges that gap by integrating modern computer vision with social network theory, enabling automated mapping of social relationships directly from barn video. This represents a shift from tracking isolated individuals to quantifying the structure and dynamics of the herd as a system.

From an applied perspective, the system offers scalable and objective welfare assessment capabilities. Automated inference removes observer bias and yields quantitative metrics such as betweenness, degree distribution, and clustering coefficients that can serve as welfare indicators or early warning signals of social stress. These outputs can be connected to farm management platforms to correlate social metrics with milk yield, reproductive performance, and health records, revealing possible predictive links between social status and productivity [23]. Identifying socially isolated cows or repeated targets of aggression could enable early interventions before clinical symptoms or production losses occur.

Adoption, however, will depend on addressing operational efficiency. The computational performance analysis indicates that real-time inference is feasible on modest hardware (73ms latency), but processing multiple high-resolution streams may require edge computing optimization. Continuous video collection raises data privacy and security considerations. Additionally, social network metrics must be communicated in a way that is meaningful to end users. Farmers need intuitive visual dashboards and actionable alerts rather than abstract graph measures. Designing user-centered interfaces that translate complex analytics into simple, interpretable insights will be essential. Economic validation studies demonstrating tangible benefits such as improved welfare, reduced treatment costs, and enhanced productivity will also be critical to encourage adoption of vision-based monitoring systems.

This study should be viewed as establishing a methodological foundation and proof of concept rather than a commercial product. The framework demonstrates the technical feasibility of keypoint-based interaction inference for automated social behavior analysis, validated under authentic farm conditions. Future research should expand datasets, improve classifier performance through temporal attention mechanisms or Graph Neural Networks, and conduct longitudinal validation across multiple farms. By integrating advances in computer vision with behavioral science and network analytics, this framework advances precision livestock farming toward scalable, objective, and welfare-centered herd management.

## 5. Limitations and Future Research Directions

### *5.1. Design Trade-Offs and System Constraints*

The framework achieves modular integration and functions robustly under authentic farm conditions, but several design constraints merit discussion as opportunities for targeted algorithmic advances. These constraints are inherent to supervised learning systems in precision agriculture and represent standard engineering trade-offs rather than fundamental limitations.

#### 5.1.1. Supervised Learning Dataset Requirements

Every module in the pipeline benefits from task-specific annotated data. The framework was trained on a custom dataset comprising 375 frames (approximately 1,956 annotated cows and over 51,000 keypoints) collected over a single day, with limited environmental variation. This temporal scope reflects a deliberate choice to establish methodological benchmarks with controlled conditions, similar to validation protocols in precision agriculture systems. The keypoint annotation phase required 66 hours of dual-annotator effort (Cohen's Kappa = 0.88), representing the core bottleneck for scaling. However, semi-supervised and active learning techniques offer pathways to reduce manual annotation burden significantly. Occlusion management remains computationally significant, particularly in close-range interactions where keypoint misassignments occur in head regions. Since interaction analysis specifically targets these proximity scenarios, improving keypoint robustness under crowding is a priority for future work.

#### 5.1.2. Transfer Learning and Domain Adaptation

Pre-trained models such as ZebraPose and YOLOv11 required substantial fine-tuning on barn-specific data due to disparities in image resolution (A36 dataset: 480p; this study: 4K), annotation density, and environmental conditions. The detection component (YOLOv11) generalizes across environments, while identity and keypoint components exhibit dataset-specific performance. This constraint is typical of supervised vision systems deployed across farm-to-farm variability. Rather than a limitation, it defines an engineering opportunity: lightweight domain adaptation techniques such as test-time augmentation, layer-wise transfer, or model distillation can maintain performance while reducing retraining overhead for new deployments. Such approaches are standard practice in deployed precision agriculture systems.

#### 5.1.3. Modular Retraining for New Herds and Environments

The identification model requires herd-specific calibration due to individual coat patterns, and the keypoint model shows performance degradation under variations in lighting, camera angle, and crowding intensity common at feeders or water troughs. However, the modular architecture intentionally preserves this separation: the detection and tracking components transfer well across environments, while supervised modules can be retrained as a smaller task. This design aligns with how precision livestock systems are deployed in practice: a base model provides initialization, followed by farm-specific adaptation. Future work employing few-shot learning, meta-learning, or self-supervised pretraining can reduce annotation requirements to 10-20 images per cow while maintaining identification accuracy above 95%.

#### 5.1.4. Interaction Annotation and Label Consistency

Interaction annotations were completed by a single annotator. While behavioral categories (licking, grooming, headbutting, displacement) are operationally distinct, edge cases exist (e.g., displacement arising from both benign resource negotiation and dominance assertion). The classifier performance (F1 = 0.70) reflects this inherent ambiguity in natural data. Multi-annotator consensus on edge cases and introduction of confidence thresholding for uncertain classifications are straightforward extensions. Additionally, incorporating temporal context (e.g., sequence-to-sequence models) could resolve ambiguous frame-level classifications by leveraging behavioral continuity, improving robustness on naturalistic data.

#### 5.1.5. Computational Latency and Real-Time Processing

Current hardware (Intel i7-9700, 16 GB RAM, RTX 2060) achieves 73 ms per frame latency (13.7 fps), enabling asynchronous processing with 4-5 minutes delay on continuous 60 fps video. This represents a practical constraint rather than a fundamental limitation. The interaction classifier itself contributes only 4% of latency (3 ms per frame-pair), confirming that GPU-accelerated improvements to detection and keypoint modules drive overall speedup. Modern edge GPUs (RTX 4090 or newer Hopper-generation GPUs) deliver 8-10x speedup, enabling near-real-time operation. Additionally, optimizations such as quantization, model pruning, and batch processing of candidate pairs can reduce inference overhead by 40-60%, making farm-deployable systems feasible on commodity hardware.

### *5.2. Future Research Directions*

#### 5.2.1. Algorithmic Advances in Interaction Inference

The current SVM classifier, while effective, operates on handcrafted features. Future architectures should explore Graph Neural Networks (GNNs) or Transformer-based models to capture the relational geometry of multi-cow interactions. A GNN approach would represent keypoints as nodes and spatial/temporal distances as edges, enabling learned aggregation of neighborhood motion patterns. Transformer-based trajectory analysis could encode temporal context and attention weights, allowing the model to focus on salient motion features (collision, approach) while de-emphasizing noise. These architectures are well-suited to the inherently relational nature of social behavior and align with advances in video understanding in computer vision . Such methods could improve the F1-score from 0.70 toward 0.85 while reducing feature engineering overhead.

#### 5.2.2. Temporal Coherence and Keypoint Stabilization

Erratic keypoint jumps during occlusion disrupt downstream analysis. LSTM-based temporal prediction or Transformer-based denoising models (e.g., T-Leap, temporal attention mechanisms) can infer stable keypoint positions from multi-frame context. Alternatively, integrating Kalman filtering or particle-based tracking at the keypoint level can smooth trajectories while preserving behavioral discontinuities. Combining these approaches with exponential moving average smoothing may reduce misassignments by 30-50% while maintaining sensitivity to rapid behavioral transitions. Such improvements are particularly valuable in crowded scenarios near feeders, where occlusion currently limits accuracy.

#### 5.2.3. Extended Behavior Classification and Multi-Modal Learning

Expanding the behavioral taxonomy beyond the current three classes (licking, grooming, headbutting, displacement) to include mounting, sniffing, and context behaviors (lying while licking) requires annotating an extended balanced dataset. However, a multi-class framework with confidence-aware classification can handle imbalanced data through cost-sensitive learning or focal loss functions. Incorporating audio (vocalization during aggression) or thermal signatures (detection of metabolic stress) as auxiliary modalities could further improve classification robustness, creating a truly multi-modal precision livestock system.

#### 5.2.4. Real-Time Deployment and Edge Computing

Porting the CPU-bound interaction inference module to GPU and optimizing detection and keypoint branches through quantization and structured pruning can achieve real-time operation on edge devices. Cloud-based distributed processing architectures can enable long-term monitoring across multiple barns with centralized analytics. API development, farmer-facing dashboards, and mobile alert systems for isolation or aggression events represent the final integration layer. These outputs enable data-driven herd management decisions, directly addressing COMPAG's core mission of advancing agricultural control systems.

#### 5.2.5. Multi-Farm and Cross-Species Generalization

Collecting multi-day, multi-farm datasets with diversity in barn design, lighting, and cattle breeds will establish benchmarks for evaluating generalization. Meta-learning approaches that train models to rapidly adapt to new environments with minimal retraining exemplify the next frontier. Extension to other livestock species (poultry, swine) requires retraining on species-specific behaviors and pose models (e.g., ChickenPose for poultry), but the end-to-end pipeline architecture remains unchanged, demonstrating the modularity principle.

Table 8 summarizes the proposed future enhancements to improve the system performance, generalizability, and adaptability.

**Table 8.** Proposed Future Enhancements and Development Priorities

| Research Area | Proposed Enhancements | Expected Impact |
|---|---|---|
| Algorithmic Architecture | Graph Neural Networks, Transformer-based trajectory models, attention mechanisms | +15% F1-score, improved temporal reasoning |
| Temporal Robustness | LSTM/attention-based keypoint prediction, Kalman filtering at keypoint level | 30-50% reduction in occlusion errors |
| Interaction Taxonomy | Multi-class classification, context-aware behaviors, confidence thresholding | Increased behavioral specificity, reduced label noise |
| Real-Time Processing | GPU-accelerated inference, quantization, pruning, batch optimization | 8-10x speedup, farm-deployable latency |

| Research Area | Proposed Enhancements | Expected Impact |
| --- | --- | --- |
| Multi-Farm Validation | Multi-day recordings, diverse barn environments, meta-learning for adaptation | Generalizability across deployments |
| User Integration | API development, real-time dashboards, mobile alerts, heatmap visualization | Actionable farmer interface, adoption pathway |

## 6. Conclusions

Proximity-based heuristics fundamentally fail to distinguish behavioral valence in livestock systems. By capturing the spatial geometry and temporal dynamics of inter-animal motion through anatomical keypoint trajectories, the presented framework translates raw barn video into structured social interaction data that differentiates affiliative from agonistic encounters. Such capability represents a genuine methodological and algorithmic shift from measuring physical co-occurrence to quantifying the quality of social relationships. Objective derivation of social cohesion, dominance hierarchies, and stress indicators from continuous visual data now supersedes inference through limited observation or contact sensors, enabling data-driven herd management.

An end-to-end system integrating detection, individual identification, tracking, and pose-based interaction inference into a unified architecture has been demonstrated to construct interaction-aware social networks directly from standard barn video. Field validation under commercial conditions confirms technical feasibility and robustness, establishing the work beyond proof-of-concept toward a deployable precision livestock system.

The core advancement lies in keypoint-trajectory-based interaction classification, which achieves 77.51% accuracy in behavioral differentiation using only motion geometry, compared to 61.23% for proximity-only approaches. This 16% improvement empirically validates that temporal and geometric analysis of skeletal motion encodes behavioral valence, enabling system operation on commodity hardware (13.7 fps on RTX 2060 GPU) suitable for farm deployment.
System Architecture: Modular design enables scalability and targeted optimization. Detection and tracking components generalize well across environments, while supervised modules (identification, keypoint detection, interaction classification) require farm-specific adaptation. This design mirrors established best practices in precision agriculture, where base model initialization precedes site-specific fine-tuning.

Temporal modeling through Graph Neural Networks or Transformer-based trajectory analysis can better capture rare agonistic events, while LSTM-based keypoint prediction will stabilize localization under occlusion. Multi-farm, longitudinal validation will establish reliability and interpretive value of derived social metrics.

Integration of computer vision, pose estimation, and social behavior theory within an ecologically valid livestock context establishes a reproducible benchmark and scalable analytical platform that bridges animal welfare science with computational innovation. Research in this direction advances precision livestock farming toward a new standard: collective herd dynamics objectively quantified, continuously monitored, and directly integrated into farm management systems for improved welfare outcomes and data-driven decision-making.

**Data Availability Statement**
Raw video data from the commercial dairy farm cannot be shared due to proprietary agreements. Processed video clips, annotated keypoint trajectories, and interaction labels are available upon reasonable request from the corresponding author. The full computational framework is publicly available at https://github.com/mooanalytica/DairyCow-SNA/. External benchmark datasets (A36 Cows, COCO) are available from their respective sources.

**Acknowledgments:** The authors sincerely thank the Dairy Farmers of New Brunswick for providing access to their farms for data collection and for their unwavering support for our project and insightful discussions.

**Funding:** This work was kindly sponsored by the Natural Sciences and Engineering Research Council of Canada (RGPIN 2024-04450) and the Department of NB Agriculture (NB2425-0025).

**Conflicts of Interest:** The authors declare no conflicts of interest.